\pdfoutput=1
\documentclass[australian,english,american,letterpaper,conference,10pt,twocolumn]{IEEEtran}
\usepackage[T1]{fontenc}
\usepackage[latin9]{inputenc}
\usepackage{babel}
\usepackage{array}
\usepackage{float}
\usepackage{amsmath}
\usepackage{amssymb}
\usepackage{graphicx}

\makeatletter

\providecommand{\tabularnewline}{\\}
\newcommand{\lyxdot}{.}

\usepackage{float}
\usepackage{times}
\usepackage{microtype}

\@ifundefined{showcaptionsetup}{}{%
 \PassOptionsToPackage{caption=false}{subfig}}
\usepackage{subfig}
\makeatother

\begin{document}
\title{Unsupervised Anomaly Detection on Temporal Multiway Data }
\author{Duc Nguyen, Phuoc Nguyen, Kien Do, Santu Rana, Sunil Gupta and Truyen
Tran\\
Applied AI Institute, Deakin University, Australia}

\maketitle
\global\long\def\Loss{\mathcal{L}}%
\global\long\def\Expect{\mathbb{E}}%
\global\long\def\MatrixNorm{\mathcal{M}}%
\global\long\def\Normal{\mathcal{N}}%
\global\long\def\Irm{\mathrm{I}}%
\global\long\def\Real{\mathbb{R}}%
\global\long\def\transpose{\mathcal{\mathrm{T}}}%
\global\long\def\Xb{\boldsymbol{X}}%

\begin{abstract}
\selectlanguage{english}%
Temporal anomaly detection looks for irregularities over space-time.
Unsupervised temporal models employed thus far typically work on sequences
of feature vectors, and much less on temporal multiway data. We focus
our investigation on two-way data, in which a data matrix is observed
at each time step. Leveraging recent advances in matrix-native recurrent
neural networks, we investigated strategies for data arrangement and
unsupervised training for temporal multiway anomaly detection. These
include compressing-decompressing, encoding-predicting, and temporal
data differencing. We conducted a comprehensive suite of experiments
to evaluate model behaviors under various settings on synthetic data,
moving digits, and ECG recordings. We found interesting phenomena
not previously reported. These include the capacity of the compact
matrix LSTM to compress noisy data near perfectly, making the strategy
of compressing-decompressing data ill-suited for anomaly detection
under the noise. Also, long sequence of vectors can be addressed directly
by matrix models that allow very long context and multiple step prediction.
Overall, the encoding-predicting strategy works very well for the
matrix LSTMs in the conducted experiments, thanks to its compactness
and better fit to the data dynamics.\selectlanguage{american}%

\end{abstract}

\section{Introduction}

\selectlanguage{australian}%

Unsupervised detection of unusual temporal signals that deviate from
the norms is vital for intelligent agents. In the most common setting,
data is represented as a sequence of 1D feature vectors, thus lacking
expressiveness over the data sequences whose elements are multiway
(2D or more). For examples, electroencephalography (EEG) spectrograms
are sequences of channel-frequency matrices; and a video can be represented
as a sequence of clips, each of which can be summarised using a covariance
matrix (e.g., see \cite{guo2013action}). Classical methods require
flattening these matrices into vectors, thus breaking the internal
structure of the data. This results in demand for more data, bigger
models and higher computational burden to compensate for the loss.
But a bigger model would be ineffective in unsupervised anomaly detection
because irregularity can sneak in as a form of overfitting, thus reducing
the discriminative power. A better way is to seek a compact model
that is native to multiway data.

We investigate the use of matrix recurrent neural networks (RNNs)
for unsupervised anomaly detection for temporal two-way data \cite{do2017learning},
as inspired by the recent successes of RNNs on temporal one-way domains
\cite{fan2019multi,hundman2018detecting,zhang2019deep}. In particular,
Matrix Long Short-Term Memory (matLSTM) is chosen for its high memory
capacity, compactness, and ease of training. matLSTM maintains a dynamic
memory matrix of past seen data, i.e., it compresses the variable-size
history tensor into a fixed-size matrix. This suggests we may use
the compression loss as a measure of abnormality. More concretely,
if a data sequence is compressible by the learnt model, we can reconstruct
the sequence with little loss. Further, as matLSTM is also able to
capture long-term dependencies between distant input matrices, we
can predict the future data after seeing a sufficiently long history.
This suggests prediction loss can be used as a measure of abnormality.

Our investigation reveals interesting phenomena not previously reported,
to the best of our knowledge. While using reconstruction loss is popular
in autoencoder-based methods, the presence of the high-capacity memory
in \emph{compact matLSTM enables the model to memorise the data noise}
\emph{in the test sequences}. This is different from the usual problem
of overfitting in large models, where the noise is memorised for the
training data only. Thus a small reconstruction loss implies neither
better model fit nor normality under noisy conditions. The predictive
model, fortunately, does not suffer from this drawback. 

Second, with prediction-based anomaly detection, we can operate on
sequences of vectors that are much longer than before. A standard
vector LSTM would have a hard time learning from $T\ge100$ steps
in the past and predicting $N\ge10$ steps ahead. This is because
the long history makes gradient flow much more difficult due to the
nonlinearity of the RNN system, and the far future would quickly accumulate
prediction errors to the point that normality cannot be judged upon.
But this is not an issue for matLSTM as we can stack $N$ vectors
into a matrix, thereby operating on a history of $\left\lfloor \frac{T}{N}\right\rfloor $
length and one step prediction, while still capturing the temporal
relationship between the vectors through matrix operations.

The main contributions of this work can be summarised as follows:
\begin{itemize}
\item We investigate the applications of matrix recurrent neural networks
for unsupervised anomaly detection for temporal multiway data.
\item Two anomaly detection settings (reconstruction and prediction) are
examined, and the empirical results on synthetic data, moving digits
and electrocardiogram (ECG) readings are reported.\selectlanguage{american}%
\end{itemize}

\section{Related Work \label{sec:Related}}

\selectlanguage{english}%
Anomaly detection (AD) in sequential data has been studied widely
in the literature, with learning methods ranging from supervised,
semi-supervised to unsupervised \cite{chalapathy2019deep}. In real-world
settings where data has no label, supervised and semi-supervised methods
are thus inapplicable. Therefore, one has to resort to unsupervised
approaches, which assume the majority of instances are normal and
a set of descriptors are learnt to represent their distribution. The
instances which are under-represented by those descriptors are deemed
anomalous. Conventionally, shallow anomaly detectors such as the One-Class
SVM (OC-SVM) or Support Vector Data Description (SVDD) are used \cite{chandola2009anomaly}.
However, these methods require substantial efforts for manual feature
engineering, making them inefficient to work on high-dimensional data.
In contrast, deep learning approaches, such as deep autoencoders (AE)
\cite{hinton2006reducing} and their variants can extract compact
features from high-dimensional inputs and leverage those for AD. 

Coupling AE with sequential models such as RNNs is one of the common
strategies to detect anomalies in sequential data \cite{malhotra2016lstm,marchi2015non,morais2019learning}.
In \cite{malhotra2016lstm}, the authors utilised the LSTM encoder-decoder
framework to compress and decompress the input sequence. Since the
model is exposed to normal samples during training, it will yield
high reconstruction error upon observing anomalous samples. Other
than reconstruction-based methods, prediction-based methods are also
proposed for this task. In \cite{marchi2015non}, an LSTM decoder
is tasked to predict a segment of audio signals given the previous
ones. While normal signals are predictable, the novel ones are not
well-predicted. Although two strategies work in certain cases, the
model can be forced to memorise the input or favour information from
recent observations over older ones. In \cite{morais2019learning},
a composite model was adopted to alleviate this behaviour. The reconstruction
branch aims to reconstruct the first half of the sequence, while the
prediction branch aims to predict the future observations. Hence,
the learnt representation by the encoder must store information for
both goals, which regularise the model and improve its generalisation
ability. 

Similar to the above approaches, our work falls into the unsupervised
and self-supervised learning branches where we use the encoder-decoder/reconstruction
\cite{kieu2019outlier,malhotra2016lstm} and sequence-to-sequence/prediction
architectures \cite{marchi2015non}. However, while their methods
operate on (sequence of) vector inputs, we target matrix structured
data that are permutation-invariant \cite{do2017learning,gao2017matrix}.
This is different from outlier detection in images \cite{sabokrou2018adversarially}
and videos \cite{morais2019learning,morais2020learning} where the
translation-invariant assumption in these data no longer holds for
permutation-invariant data.

Closely related to our work is \cite{do2017learning}, in which matrix-structured
LSTM was proposed, but not for anomaly detection. Matrix-structured
latent variable model has been studied in \cite{wang2019spatial},
where a matrix normal distribution was used for both the prior and
posterior in the latent space. However, although the latent is a matrix,
this work used the encoder and decoder that required vectorising its
input and output. It was shown that vectorised inputs and outputs
will break the permutation invariant properties \cite{hartford2018deep}.
In contrast, our models maintain the matrix structure in the hidden
states as well as using matrix network for the mappings, thus is more
compact. 

\selectlanguage{american}%

\section{Methods \label{sec:Method}}

\selectlanguage{australian}%
Temporal two-way data is a sequence $\Xb_{1:T}=\left(X_{1},X_{2},\ldots,X_{T}\right)$
where the input at each step $X_{t}\in\Real^{n_{r}\times n_{c}}$
is a matrix representing data observed at time $t$. Without loss
of generality, let $n_{r}\ge n_{c}$. We focus on the setting where
\emph{the matrix rows (or columns) can be permuted} without changing
the matrix's key characteristics. This makes the setting applicable
to a wider range of application scenarios.

Standard recurrent neural networks (RNN, LSTM and GRU) assume vectorised
data, and thus have space complexity of $\mathcal{O}\left(n_{r}n_{c}k\right)$,
where $k$ is the size of the hidden layer. For a typical model with
rich representation, $k$ is in the order of $n_{r}n_{c}$, i.e.,
resulting in the space complexity of $\mathcal{O}\left(n_{r}^{2}n_{c}^{2}\right)$.
The model size explodes for large $n_{r}$ and $n_{c}$. Hence, a
more compact modelling is needed.

\subsection{Matrix LSTM}

Matrix LSTM (matLSTM) \cite{do2017learning} is an extension of the
LSTM designed to effectively deal with sequences of matrices. Like
LSTM, the matLSTM maintains a short-term memory that summarises the
historical data. However, unlike LSTM, the matLSTM uses matrices to
natively represent input $X_{t}$, neural activations in hidden state
$H_{t}\in\Real^{k_{r}\times k_{c}}$ and \emph{working memory} $C_{t}\in\Real^{k_{r}\times k_{c}}$.
matLSTM compresses the data tensor $\Xb_{1:t}$ into $C_{t}$. It
is also highly compact as the number of parameters typically scales
linearly with $n_{r}^{2}$.

Let us define the following operation:
\begin{align*}
\text{mat}\left(X,H;\theta\right) & =U_{xh}^{\top}XV_{xh}+U_{hh}^{\top}HV_{hh}+B
\end{align*}
where $U_{xh}\in\mathbb{R}^{k_{r}\times n_{r}}$, $V_{xh}\in\mathbb{R}^{n_{c}\times k_{c}}$,
$U_{hh}\in\mathbb{R}^{k_{r}\times k_{r}}$, $V_{hh}\in\mathbb{R}^{k_{c}\times k_{c}}$,
and $B\in\mathbb{R}^{k_{r}\times k_{c}}$ are free parameters. Upon
seeing a new input, the memory is refreshed using
\[
C_{t}=F_{t}\odot C_{t-1}+I_{t}\odot\hat{C}_{t}
\]
where $\hat{C}_{t}=\text{tanh}\left(\text{mat}\left(X_{t},H_{t-1};\theta_{c}\right)\right)$,
and $I_{t},F_{t}\in(\boldsymbol{0},\boldsymbol{1})$ are input and
forget gates. The state is estimated using $H_{t}=O_{t}\odot C_{t}$,
where $O_{t}\in(\boldsymbol{0},\boldsymbol{1})$ is output gate. The
gates are computed as $I_{t}=\sigma\left(\text{mat}\left(X_{t},H_{t};\theta_{i}\right)\right)$,
$F_{t}=\sigma\left(\text{mat}\left(X_{t},H_{t};\theta_{f}\right)\right)$
and $O_{t}=\sigma\left(\text{mat}\left(X_{t},H_{t};\theta_{o}\right)\right)$,
respectively, for $\sigma\left(\cdot\right)$ is element-wise logistic
function.

Prediction at time $t$ takes $H_{t}$ as input and computes a matrix
feedforward net:
\[
\hat{Y}_{t}=\text{matnet}\left(H_{t}\right)
\]
whose basic fully connected layers assume the form: $Z^{l+1}=f\left(U^{l}Z^{l}V^{l}+B^{l}\right)$
for element-wise nonlinear transformation $f$.

Training proceeds by minimising a loss function. For example, for
continuous outputs, we may use the quadratic loss: $L_{mse}=\frac{1}{T}\sum_{t=1}^{T}\left\Vert Y_{t}-\hat{Y}_{t}\right\Vert _{F}^{2}$,
where $\left\Vert \cdot\right\Vert _{F}$ is Frobenius norm. For binary
outputs, a cross-entropy loss is applied: $L_{cross}=$

\begin{multline*}
-\frac{1}{T}\sum_{t=1}^{T}\left\Vert Y_{t}\odot\log\sigma\left(\hat{Y}_{t}\right)+\left(1-Y_{t}\right)\odot\log\left(1-\sigma\left(\hat{Y}_{t}\right)\right)\right\Vert _{1}
\end{multline*}

\subsection{Lossy sequence compression \label{subsec:Lossy-sequence-compression}}

As the memory matrix  $C_{T}$ at time step $T$ of the matLSTM is
a lossy compression of the data $\Xb_{1:T}$, we can use the reconstruction
loss as a measure how regular the sequence is, similar to the case
of vector sequence \cite{malhotra2016lstm}. That is because an abnormal
sequence does not exhibit the regularities, it is hardly compressible,
and thus its reconstruction error is expected to be higher than the
error in the normal cases.

\begin{figure}[H]
\centering{}\includegraphics[height=3.5cm]{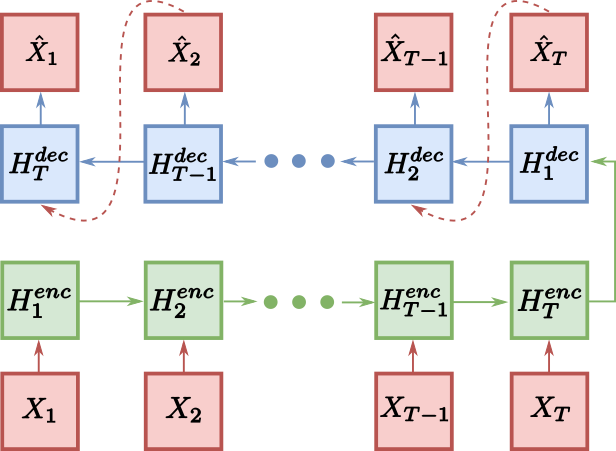}\caption{(Matrix) LSTM AutoEncoder model \label{fig:autoencoder}}
\end{figure}

The model thus has two components: an \emph{encoder} matLSTM which
compresses $\Xb_{1:T}$ into $C_{T}^{enc}$ (and $H_{T}^{enc})$ by
reading one matrix at a time; and a \emph{decoder} matLSTM decompresses
the memory into $\hat{\Xb}_{T+1:2T}$ by predicting one matrix at
a time. The decoding takes the initial state $H_{T}^{enc}$ and proceeds
backward, starting from the last element, until the first. At each
step $t=T+1,T+2,...,2T$, it predicts an output matrix $\hat{X}_{t}$.
See Fig.~\ref{fig:autoencoder} for a graphical illustration. We
denote the two models employing this strategy as LSTM-AutoEncoder
and matLSTM-AutoEncoder.

The anomaly score $e_{\boldsymbol{X}}$ is computed as the mean reconstruction
error $\frac{1}{T}\sum_{t=1}^{T}\left\Vert X_{t}-\hat{X}_{2T-t+1}\right\Vert _{F}^{2}$
for continuous data, and mean cross-entropy for binary data. 

\subsection{Predicting the unrolling of sequence \label{subsec:Predicting-the-unrolling}}

An alternative to the autoencoder method relies on the premise that
if a sequence is regular (i.e., normal), the history may contain sufficient
information to predict several steps ahead, as the temporal regularities
unrolled over time. This is arguably based on a stronger assumption
than the compress-decompress strategy in the autoencoder, because
we cannot rely on the working memory $C_{T}$ of the current past
only but also the statistical regularities embedded in model parameters.

\begin{figure}[H]
\begin{centering}
\includegraphics[height=3.5cm]{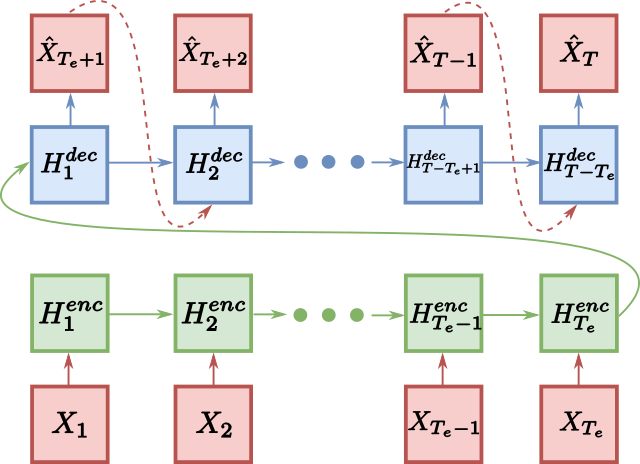}
\par\end{centering}
\caption{(Matrix) LSTM Encoder-Predictor predictive model \label{fig:enc_dec_predict}}
\end{figure}

More formally, given a past sub-sequence $\Xb_{1:T_{e}}$ we want
to predict the future sub-sequence $\Xb_{T_{e}+1:T}$ using $P\left(\Xb_{T_{e}+1:T}\mid\Xb_{1:T_{e}}\right)$.
At time $t\le T_{e}$, the encoder reads the past matrices into memory.
At time $t>T_{e}$, the decoder predicts future matrices, one at a
time. Fig.~\ref{fig:enc_dec_predict} illustrates the encoder-predictor
architecture. We denote the two models employing this strategy as
LSTM-Encoder-Predictor and matLSTM-Encoder-Predictor.

The anomaly score $e_{\boldsymbol{X}}$ is computed as the mean prediction
error $\frac{1}{T-T_{e}}\sum_{t=T_{e}+1}^{T}\left\Vert X_{t}-\hat{X}_{t}\right\Vert _{F}^{2}$
for continuous data, and mean cross-entropy for binary data. 

\subsection{Stacking LSTM layers \label{subsec:Stacking-LSTM-layers}}

Raw data may be too noisy to provide informative outliers signals.
This necessitates data denoising or abstraction. We propose to abstract
data by using the lower LSTM states as input for the higher LSTM.
That is, the stack of LSTMs is trained on a layerwise manner, starting
from the bottom to the top. At each layer, we compute an outlier score,
using methods in Sec.~\ref{subsec:Lossy-sequence-compression} and
Sec.~\ref{subsec:Predicting-the-unrolling}. How to combine the scores
across layers remains open. For simplicity, we use the score at the
top layer.

\subsection{Dynamics of changes \label{subsec:Dynamics-of-changes}}

If $X_{t}$ is an image, one may argue that since the loss and anomaly
score are based on pixel intensities, the models focus too much on
the appearance of the digits, and less on its dynamics. Thus the detection
of dynamic abnormality may be less effective as a result. To test
whether it is the case, we also study the dynamics of changes, that
is, instead of studying the original sequence $\left\{ X_{t}|1\le t\le T\right\} $,
we study the differences between time step, that is, $\left\{ \Delta X_{t}=X_{t+1}-X_{t}|1\le t<T\right\} $.\selectlanguage{american}%

\section{Experiments and Results}

\selectlanguage{australian}%
We experimentally validate our proposed strategies of using matLSTM
for unsupervised temporal two-way anomaly detection on three datasets:
(1) synthetic sequences of binary matrices, (2) sequences of moving
digits and (3) ECG recordings.

\textbf{Experimental setup:} In all experiments, we compare the performances
among LSTM, matLSTM and 3D-CNN models. The 3D-CNN models can handle
spatio-temporal input in which two of the dimensions are for space
and the remaining dimension is for time. We adopt two variations of
the 3D-CNN to match the LSTM counterparts. The first, named 3D-ConvAE,
follows compress-decompress strategy is. The second, named 3D-Conv-Predictor,
encodes the past observations and predict future observations. For
predictive models, we empirically use the first half of the input
sequence as context to predict the second half. For LSTM models,
which require vector inputs, we flatten the matrices into vectors.
For LSTM and matLSTM models, in both compression and prediction strategies,
we use a \emph{conditional} decoder which, at each decoding step,
uses the decoded output at the previous timestep as its input. We
also experimented with \emph{unconditional} decoder but found that
it was empirically worse, thus we only report results for the\emph{
}conditional decoder here. For LSTM and matLSTM models, we investigate
the ones with a single hidden layer and two hidden layers, c.f. Fig.~\ref{fig:autoencoder}
and Fig.~\ref{fig:enc_dec_predict}. We use the Adam optimiser with
a learning rate of $3\times\text{10}^{-4}$, minibatch size $64$,
and a maximum of 100 training epochs or when the model converges.
Sequence-level Area Under ROC Curve (ROC-AUC) and $F1$ performance
measures are used throughout the experiments. We repeat all experiments
3 times and report the mean and standard deviation of measures on
left-out test sets. 

\subsection{Synthetic data}

\begin{figure*}[t]
\begin{centering}
\begin{tabular}{cc}
\subfloat[Noise-free $10\times10$ input matrix. \label{fig:synthetic_small_wo_noise}]{\centering{}\includegraphics[width=0.95\columnwidth]{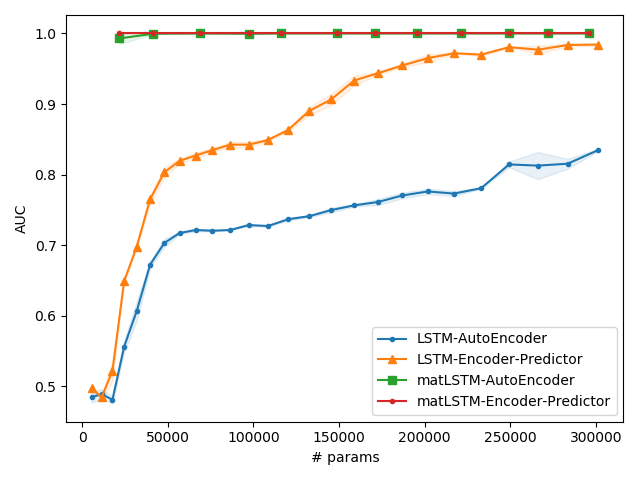}} & \subfloat[Noisy $10\times10$ matrix inputs. \label{fig:synthetic_small_with_noise}]{\centering{}\includegraphics[width=0.95\columnwidth]{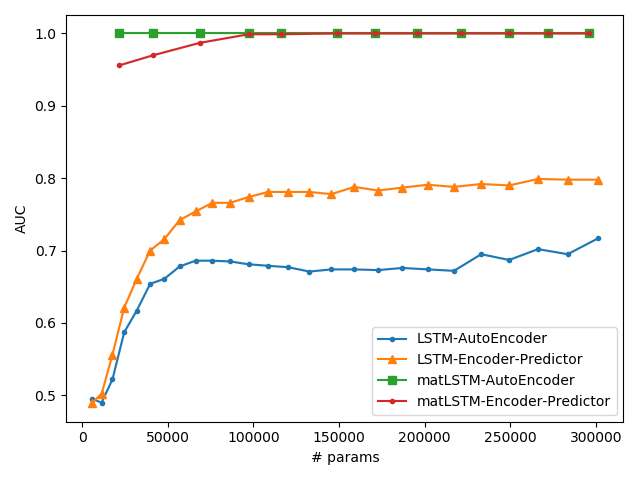}}\tabularnewline
\end{tabular}
\par\end{centering}
\caption{Behaviors of LSTM vs. matLSTM in synthetic data experiments on clean
and noisy data. Performance measure is AUC against the number of parameters.}
\end{figure*}

In this section, we conduct three different ablation studies on a
synthetic dataset: (1) noise-free, (2) noisy, and (3) long movement
pattern. In each study, the data are sequences of binary matrices
generated as follows:

\begin{align*}
X_{1} & =\left[X_{1}^{ij}\sim\text{Bern}\left(0.5\right)\right]_{i=1:n_{r},j=1:n_{c}}\\
X_{t} & =\text{circshift}\left(X_{t-1},\text{Uniform}\{1,5\}\right),\qquad t=2:T
\end{align*}
where circshift is the circular shift of the matrix to the right by
$c\sim\text{Uniform}\left\{ 1,5\right\} $ columns, $n_{r}=10$ and
$n_{c}=10$ are the number of rows and columns respectively, and $T=20$
is the sequence length. The abnormal data sequences are created similarly
but instead of the right-shifting pattern, we use a random permutation
of columns. Each of the train and test set consists of 5,000 sequences
with 5\% outlier. We randomly take 20\% of the train set for validation.
For simplicity, we only examine the single-layer models in this setting.

\subsubsection*{Noise-free data}

Fig.~\ref{fig:synthetic_small_wo_noise} compares the performance
of matLSTM to LSTM on the test set. As shown, performance improves
with more model capacity, but the matLSTMs improve much faster, suggesting
matLSTMs can capture the multi-way structure better than LSTMs. In
this setting, both compression (Sec.~\ref{subsec:Lossy-sequence-compression})
and prediction strategies (Sec.~\ref{subsec:Predicting-the-unrolling})
work well for matLSTM models while for vector models, the prediction
strategy is more robust than the compression strategy.

\subsubsection*{Noisy data}

In this setting, we randomly set 20\% of the pixels to zero at each
timestep but keep the moving pattern the same as the noise-free cases.
The results are shown in Fig.~\ref{fig:synthetic_small_with_noise}.
With noisy inputs, the performance of vector models drop, while matrix
models retain their near-perfect performance.

\subsubsection*{Longer movement pattern data}

In this setting, we investigate the changes in performances of matLSTM
vs. LSTM when the moving pattern is longer. We make the data matrices
horizontally longer by increasing $n_{c}=100$. Table~\ref{tab:synthetic-longer-pattern}
shows that both vector LSTM methods require more number of free parameters
than matLSTM to recognise the moving patterns. 

\begin{table}[h]
\begin{centering}
\begin{tabular}{|c|c|c|}
\hline 
Models & \#params & AUC\tabularnewline
\hline 
\hline 
1-layer LSTM-AutoEncoder & 472k & 94.7 $\pm$ 0.2\tabularnewline
\hline 
2-layer LSTM-AutoEncoder & 505k & 93.9 $\pm$ 0.7\tabularnewline
\hline 
1-layer LSTM-Encoder-Predictor & 472k & 95.3 $\pm$ 0.4\tabularnewline
\hline 
2-layer LSTM-Encoder-Predictor & 505k & 87.1 $\pm$ 0.5\tabularnewline
\hline 
1-layer matLSTM-AutoEncoder & 104k & 100.0 $\pm$ 0.0\tabularnewline
\hline 
2-layer matLSTM-AutoEncoder & 177k & 100.0 $\pm$ 0.0\tabularnewline
\hline 
1-layer matLSTM-Encoder-Predictor & 104k & 100.0 $\pm$ 0.0\tabularnewline
\hline 
2-layer matLSTM-Encoder-Predictor & 177k & 100.0 $\pm$ 0.0\tabularnewline
\hline 
\end{tabular}
\par\end{centering}
\caption{LSTM vs. matLSTM in longer horizontal movement pattern, matrix size
$10\times100$, on synthetic data. \label{tab:synthetic-longer-pattern}}
\end{table}

\subsection{Moving permuted digits}

\begin{figure*}[t]
\begin{centering}
\begin{tabular}{cc}
\subfloat[Noise-free input. \label{fig:mnist_wo_noise}]{\centering{}\includegraphics[width=0.95\columnwidth]{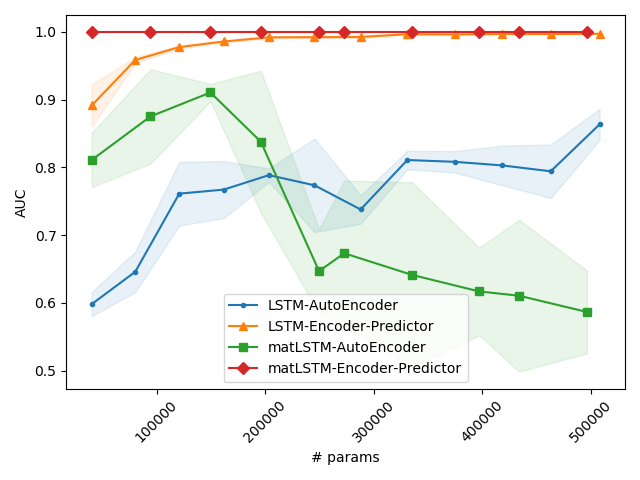}} & \subfloat[Salt and pepper noise added. \label{fig:mnist_with_noise}]{\centering{}\includegraphics[width=0.95\columnwidth]{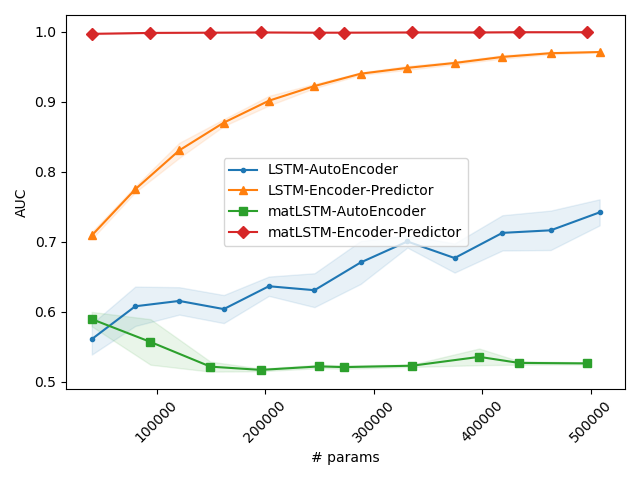}}\tabularnewline
\end{tabular}
\par\end{centering}
\caption{Results on moving MNIST.  }
\end{figure*}

\begin{figure*}[t]
\begin{centering}
\begin{tabular}{cc}
\textbf{Normal data} & \textbf{Abnormal data}\tabularnewline
\subfloat[LSTM-AutoEncoder vs. matLSTM-AutoEncoder. \label{fig:mnist_recon_normal}]{\centering{}\includegraphics[clip,width=0.8\columnwidth]{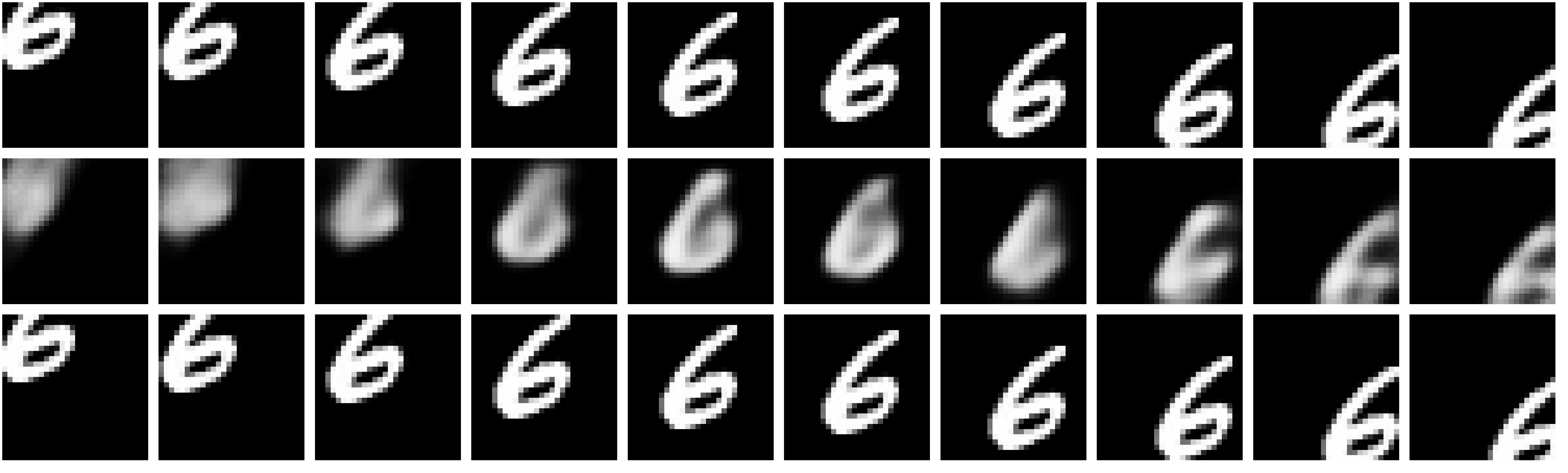}} & \subfloat[LSTM-AutoEncoder vs. matLSTM-AutoEncoder. \label{fig:mnist_recon_abnormal}]{\centering{}\includegraphics[width=0.8\columnwidth]{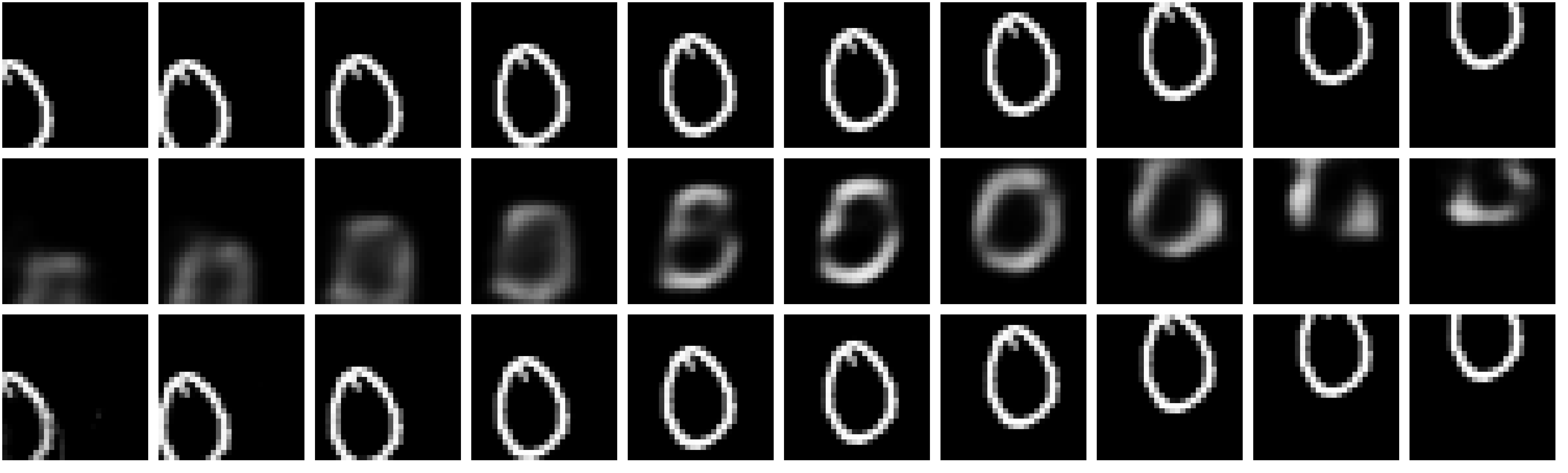}}\tabularnewline
\subfloat[LSTM-Encoder-Predictor vs. matLSTM-Encoder-Predictor. \label{fig:mnist_predict_normal}]{\centering{}\includegraphics[width=0.8\columnwidth]{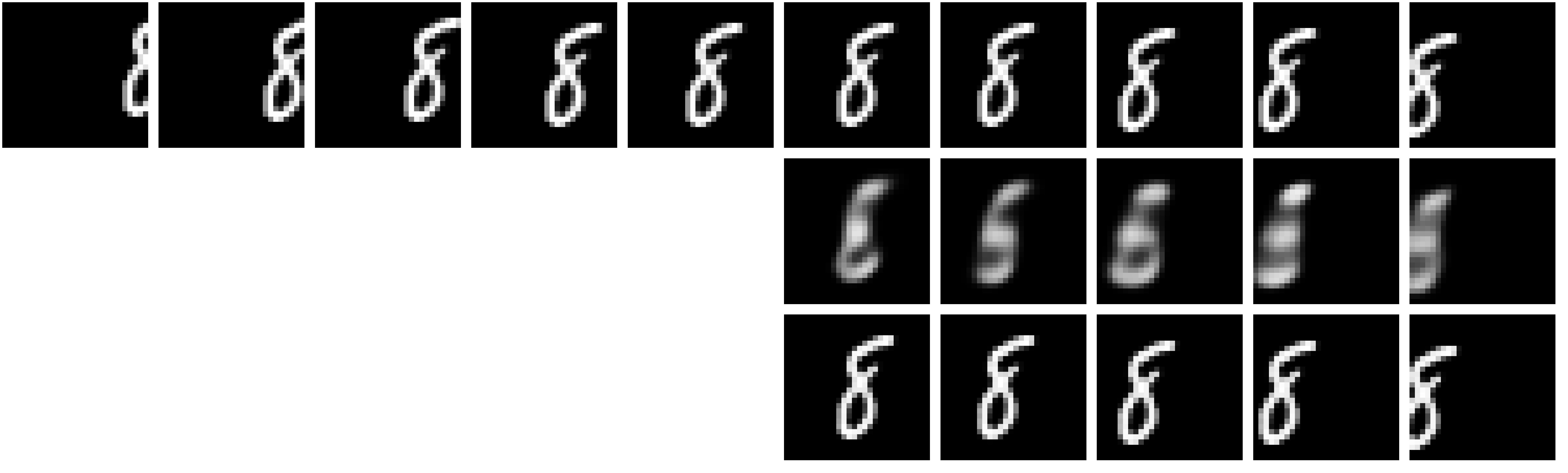}} & \subfloat[LSTM-Encoder-Predictor vs. matLSTM-Encoder-Predictor. \label{fig:mnist_predict_abnormal}]{\centering{}\includegraphics[width=0.8\columnwidth]{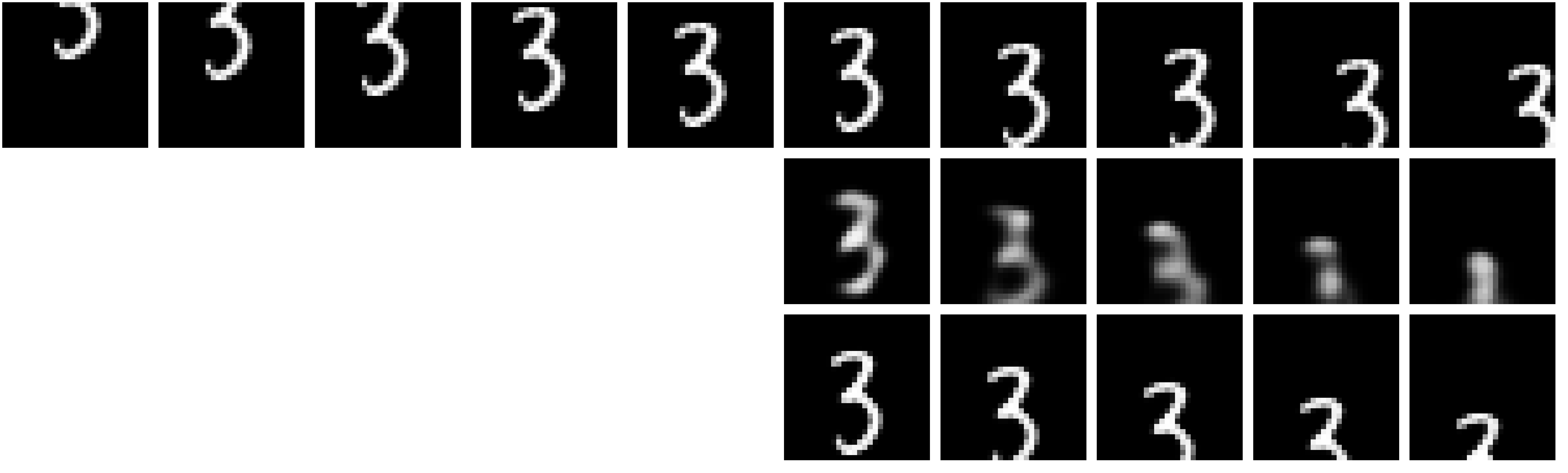}}\tabularnewline
\end{tabular}
\par\end{centering}
\caption{LSTM vs. matLSTM methods on \emph{clean} normal and abnormal data.
In each figure, the top row is the original sequence, the middle
row result is by LSTM, and the bottom row result is by matLSTM. \emph{The
rows and columns of each image are permuted in the actual models}.
Original images are shown for ease of interpretation only.}
\end{figure*}

\begin{figure*}[h]
\begin{centering}
\begin{tabular}{cc}
\textbf{Normal data} & \textbf{Abnormal data}\tabularnewline
\subfloat[LSTM-AutoEncoder vs. matLSTM-AutoEncoder. \label{fig:mnist_recon_normal_sp_noise}]{\centering{}\includegraphics[width=0.8\columnwidth]{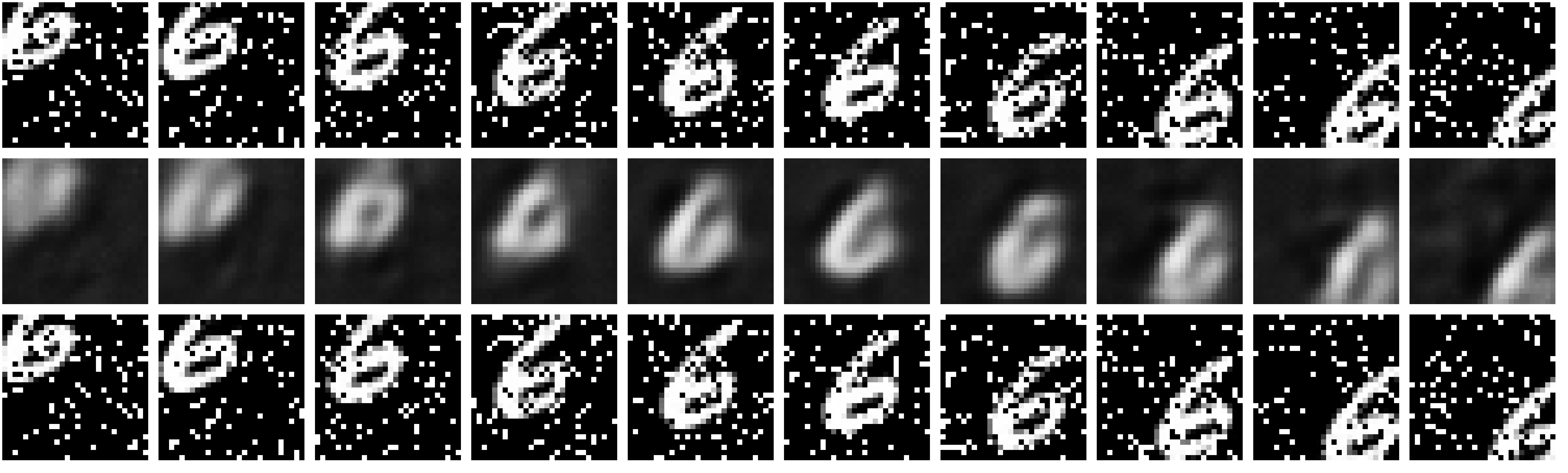}} & \subfloat[LSTM-AutoEncoder vs. matLSTM-AutoEncoder. \label{fig:mnist_recon_abnormal_sp_noise}]{\centering{}\includegraphics[width=0.8\columnwidth]{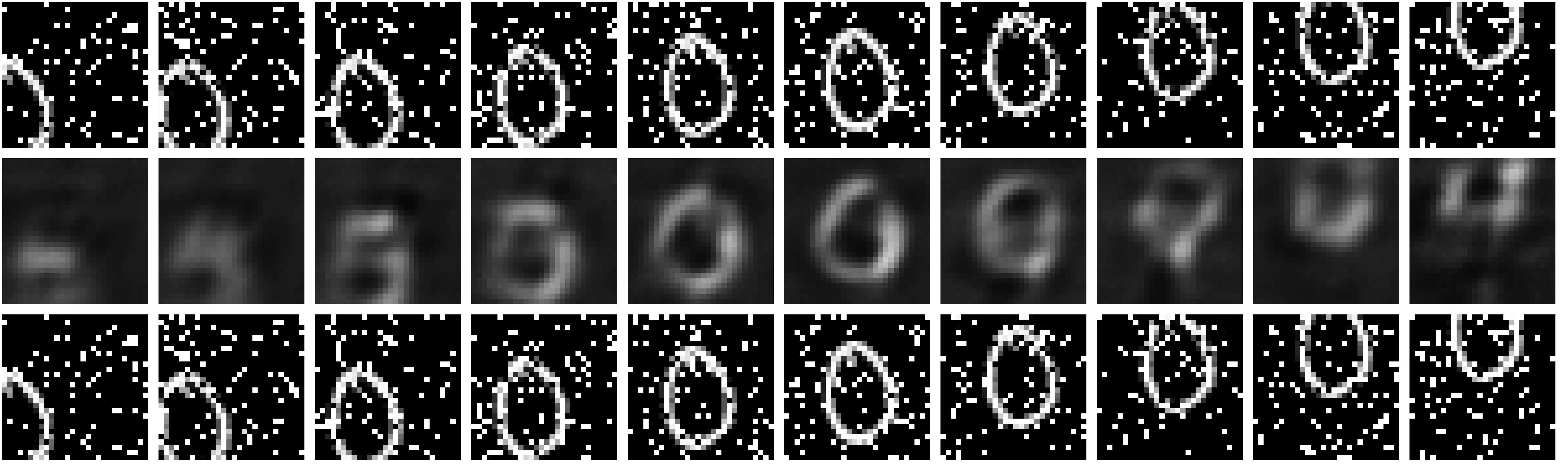}}\tabularnewline
\subfloat[LSTM-Encoder-Predictor vs. matLSTM-Encoder-Predictor. \label{fig:mnist_predict_normal_sp_noise}]{\centering{}\includegraphics[width=0.8\columnwidth]{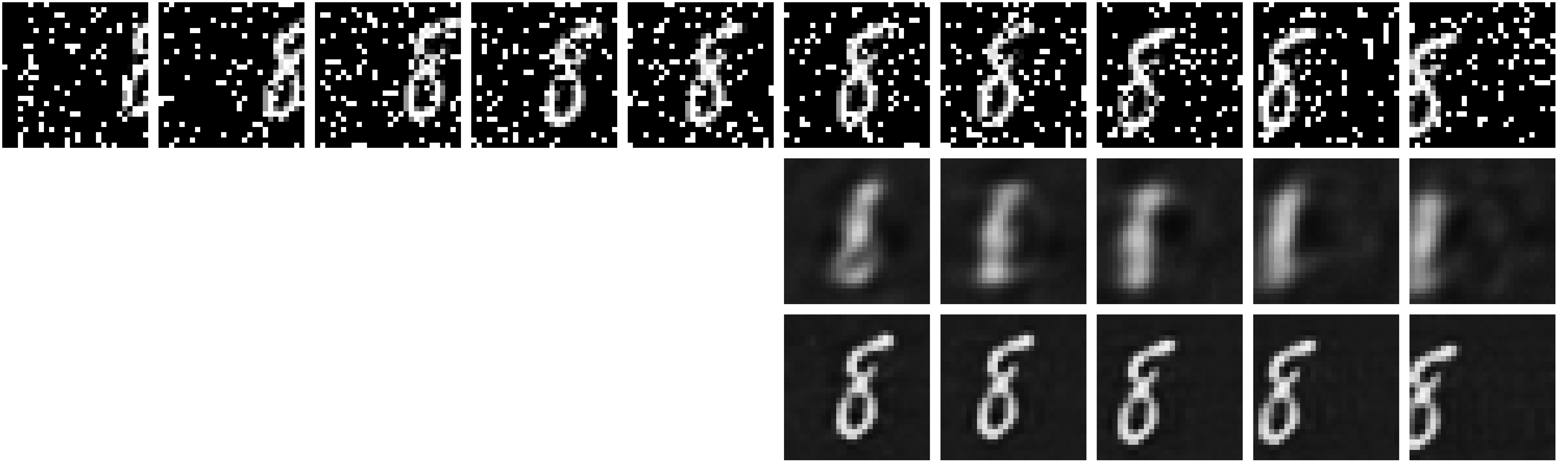}} & \subfloat[LSTM-Encoder-Predictor vs. matLSTM-Encoder-Predictor. \label{fig:mnist_predict_abnormal_sp_noise}]{\centering{}\includegraphics[width=0.8\columnwidth]{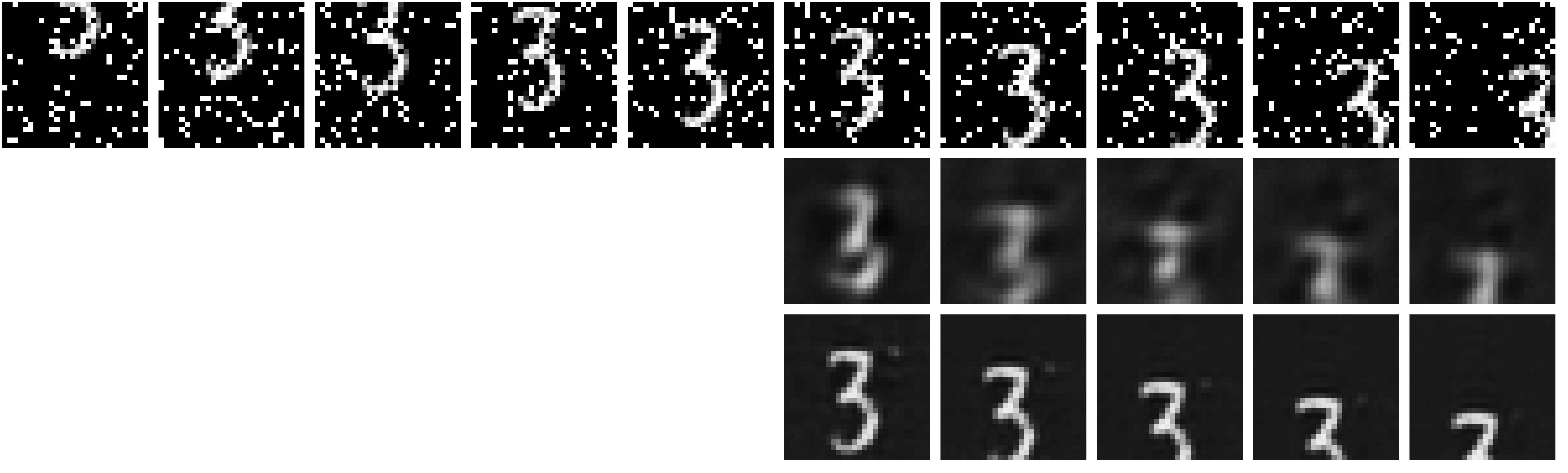}}\tabularnewline
\end{tabular}
\par\end{centering}
\caption{LSTM vs. matLSTM methods on \emph{noisy} normal and abnormal data.
In each figure, the top row is the original sequence, the middle
row result is by LSTM, and the bottom row result is by matLSTM. \emph{The
rows and columns of each image are permuted in the actual models}.
Original images are shown for ease of interpretation only. }
\end{figure*}

For ease of visualisation, we generate moving digits from the MNIST
dataset. MNIST dataset contains 60,000 images in the train set and
10,000 images in the test set. For normal sequences, we first select
a random slope between $\left[0,2\pi\right]$ and move the digit along
a straight line; for abnormal sequences, we also select a random slope
between $\left[0,2\pi\right]$ but the digit moves along a curved
line. A curved line has a slightly different moving pattern compared
to a straight line, therefore we want to investigate if all models
can discriminate between two moving patterns. We use all of the 60,000
digits in the MNIST train set to form the train set of moving digit
data, and use all the digits in MNIST test set to generate test set
of moving digit data. The train set contains only normal sequences
and the test set has 5\% outlier. 

\emph{We treat the images as matrices of pixels, whose rows or columns
are permutable}, ignoring the strict grid structure which typically
warrants CNN models. Our goal is not to compete against CNN-based
techniques, but to expose regularities found in 2D motions. To simulate
the permutation-invariant scenario, in each sequence, we permute the
rows and columns for each image following the same permutation matrix. 

The cross-entropy loss is used as the pixel intensities fall within
the range $[0,1]$. For each model, the hidden size of LSTM unit is
chosen so that the number of parameters increases from a small number
up to approximately 500K parameters.

\subsubsection{Noise-free data}

Fig.~\ref{fig:mnist_wo_noise} compares the AUC against the number
of parameters for the single-layer models. Again we observe that matrix
neural network can capture movement pattern and predict future frames
well, yielding high performance using much fewer parameters than vector
counterparts. 

We visualise reconstructed frames of the autoencoder models for a
normal and abnormal sequence in Fig.~\ref{fig:mnist_recon_normal}
and \ref{fig:mnist_recon_abnormal} respectively. We choose the LSTM-AutoEncoder
models which have the highest number of parameters from Fig.~\ref{fig:mnist_wo_noise}.
In Fig.~\ref{fig:mnist_recon_abnormal}, LSTM-AutoEncoder fails to
reconstruct the moving pattern of the sequence (the digit lies lower
in the first few frames), as they try to reconstruct the moving pattern
in a straight line. In contrast, matLSTM-AutoEncoder reconstructs
the sequence well, thus making it unable to discriminate between two
patterns and yields random performance in Fig.~\ref{fig:mnist_wo_noise}.

Similarly, we show that matLSTM model is able to encode and predict
digit movement well, also produce sharper images compared to the vector
counterpart, in Fig.~\ref{fig:mnist_predict_normal} and Fig.~\ref{fig:mnist_predict_abnormal}.
Since matLSTM is able to encode the past movement well, its prediction
will deviate from ground truth if the movement pattern is anomalous,
hence giving high prediction error. In Fig.~\ref{fig:mnist_predict_abnormal},
the digit is predicted to keep moving downward, while ground truth
frames show that the digit moves further to the right.

\subsubsection*{Adding outlier to training data}

We add 5\% of outlier into the training data, in the same way as the
testing data, to reflect a real situation where our training data
is not lean but contaminated with unknown outliers by a small proportion.
Table~\ref{tab:mnist-results-unsup-permute} shows the results for
all models, including 3D-CNN and two-layer LSTM models. With the presence
of outlier, the performances of both vector models drop. 

\begin{table}[h]
\begin{centering}
\begin{tabular}{|c|c|c|}
\hline 
Models & \#Params & AUC\tabularnewline
\hline 
\hline 
3D-ConvAE & 140k & 59.6 $\pm$ 0.1\tabularnewline
3D-Conv-Predictor & 66k & 81.9 $\pm$ 0.4\tabularnewline
\hline 
1-layer LSTM-AutoEncoder & 331k & 61.6 $\pm$ 0.7\tabularnewline
2-layer LSTM-AutoEncoder & 357k & 65.9 $\pm$ 0.4\tabularnewline
\hline 
1-layer LSTM-Encoder-Predictor & 331k & 76.8 $\pm$ 0.1\tabularnewline
2-layer LSTM-Encoder-Predictor & 357k & 75.3 $\pm$ 0.5\tabularnewline
\hline 
1-layer matLSTM-AutoEncoder & 52k & 64.9 $\pm$ 2.7\tabularnewline
2-layer matLSTM-AutoEncoder & 101k & 66.7 $\pm$ 0.6\tabularnewline
\hline 
1-layer matLSTM-Encoder-Predictor & 52k & \textbf{83.4 $\pm$ 1.7}\tabularnewline
2-layer matLSTM-Encoder-Predictor & 101k & 78.8 $\pm$ 2.2\tabularnewline
\hline 
\end{tabular}
\par\end{centering}
\caption{Moving permuted digits: Models performance (AUC) with outliers in
training data. \label{tab:mnist-results-unsup-permute}}

\selectlanguage{american}%
\selectlanguage{american}%
\end{table}

\subsubsection*{Dynamics only}

\begin{figure}[h]
\begin{centering}
\includegraphics[width=0.95\columnwidth]{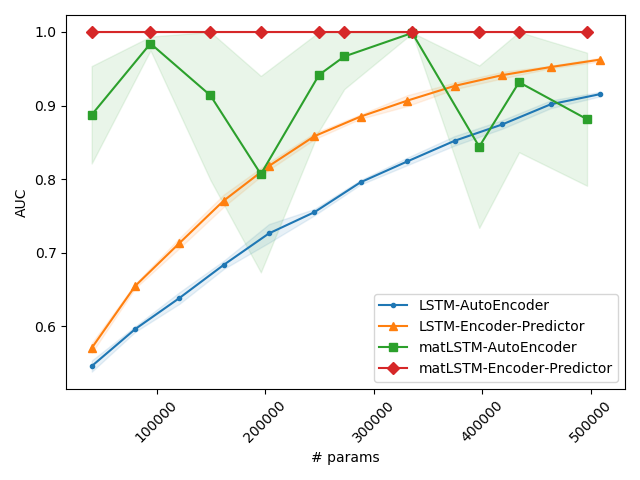}
\par\end{centering}
\caption{AUC vs. number of parameters for \emph{difference of moving permuted
digits}. \label{fig:mnist_wo_noise_input_diff} }
\end{figure}

In this experiment, we run single-layer models that take the temporal
difference $\Delta X_{t}$ as input, as described in Section~\ref{subsec:Dynamics-of-changes}.
The results are shown in Fig.~\ref{fig:mnist_wo_noise_input_diff}.
While the matLSTM-Encoder-Predictor model stills perform well, LSTM-Encoder-Predictor
model suffers a loss in performance. This suggests that while using
temporal difference as inputs may suffice given a suitable model (e.g.,
the predictive matLSTM in this case), appearance-based data can be
used to detect irregularities in trajectory dynamics.

\subsubsection{Noisy pixels}

From the noise-free data, we add salt and pepper noise to each frame
in the sequence with a probability of $0.1$ for each type of noise.
Performance vs model size for single-layer models is shown in Fig.~\ref{fig:mnist_with_noise}.
 All models require more parameters to reach high performance compared
to the noise-free counterparts. In fact, except for the prediction
strategy in matLSTM, other models struggle to learn regularities at
all. The matLSTM-AutoEncoder does not improve with more parameters.
This suggests that it may encode noise into its memory and thus cannot
discriminate between normal and abnormal sequence.

To examine the model behaviours in more detail, we visualise reconstructed
frames of the autoencoder models for normal and abnormal sequences
in Fig.~\ref{fig:mnist_recon_normal_sp_noise} and Fig.~\ref{fig:mnist_recon_abnormal_sp_noise},
respectively. The models with highest number of parameters from Fig.~\ref{fig:mnist_with_noise}
are chosen for evaluation. The figures further confirm that the matLSTM-AutoEncoder
actually memorises the noise as well as the signals, and thus cancels
out its discriminative capability. However, this is not the case for
the prediction strategy, as shown in Fig.~\ref{fig:mnist_predict_normal_sp_noise}
and Fig.~\ref{fig:mnist_predict_abnormal_sp_noise}. This is expected
because the noise is random, and there is no structure to be learnt,
and thus future prediction is still smooth. 

\subsubsection{Noisy trajectories}

A new dataset is created by randomly shifting the digits from the
original trajectories within the 3-pixel margin. In both of the train
and test set, 5\% of the sequences are abnormal. Each sequence is
then permuted using the same procedure as in the previous sections.
The performance of all models are reported in Table~\ref{tab:AUC-for-permuted-moving-digit-noisy-traj}.
We tune the number of free parameters independently for each model
using the validation set. Again the matLSTM-AutoEncoder suffers greatly
from the noise, suggesting that it does not suit for this task. The
predictive counterpart, however, performs very well, and matLSTM-Encoder-Predictor
achieves similar performance compared to LSTM-Encoder-Predictor, using
much fewer number of free parameters. The two-layer predictive models
experience overfitting, thus their performances are worse than single-layer
models.

\begin{table}[h]
\begin{centering}
\begin{tabular}{|c|c|c|}
\hline 
Models & \#Params & AUC\tabularnewline
\hline 
\hline 
3D-Conv-AE & 140k & 55.3 $\pm$ 0.0\tabularnewline
\hline 
3D-Conv-Predictor & 66k & 74.9 $\pm$ 0.3\tabularnewline
\hline 
1-layer LSTM-AutoEncoder & 331k & 60.3 $\pm$ 0.7\tabularnewline
\hline 
2-layer LSTM-AutoEncoder & 357k & 67.5 $\pm$ 0.1\tabularnewline
\hline 
1-layer LSTM-Encoder-Predictor & 331k & 82.9 $\pm$ 0.7\tabularnewline
\hline 
2-layer LSTM-Encoder-Predictor & 357k & 82.6 $\pm$ 0.2\tabularnewline
\hline 
1-layer matLSTM-AutoEncoder & 52k & 53.1 $\pm$ 0.9\tabularnewline
\hline 
2-layer matLSTM-AutoEncoder & 101k & 52.2 $\pm$ 0.2\tabularnewline
\hline 
1-layer matLSTM-Encoder-Predictor & 52k & \textbf{85.1 $\pm$ 1.2}\tabularnewline
\hline 
2-layer matLSTM-Encoder-Predictor & 101k & 82.1 $\pm$ 0.2\tabularnewline
\hline 
\end{tabular}
\par\end{centering}
\caption{AUC for moving permuted digit data with noisy trajectories. \label{tab:AUC-for-permuted-moving-digit-noisy-traj}}

\selectlanguage{american}%
\selectlanguage{american}%
\end{table}

\subsection{ECG anomalies}

We use MIT-BIH Arrhythmia dataset\footnote{https://physionet.org/content/mitdb/1.0.0/}
which contains 48 half-hour recordings of two-channel ECG signals,
obtained from 47 subjects. The recordings are digitised at 360 samples
per second. According to \cite{liu2019arrhythmias}, each heartbeat
is classified into one of five classes and detail statistics is shown
in Table~\ref{tab:ECG-heartbeats-statistics}.

\begin{table}[h]
\begin{centering}
\begin{tabular}{|l|r|}
\hline 
Classes & \# heartbeats\tabularnewline
\hline 
\hline 
Normal (N) & 90,631\tabularnewline
\hline 
Premature ventricular contraction (V) & 2,779\tabularnewline
\hline 
Supraventricular premature beat (S) & 7,236\tabularnewline
\hline 
Fusion (F) & 803\tabularnewline
\hline 
Unknown (Q) & 8,043\tabularnewline
\hline 
\end{tabular}
\par\end{centering}
\caption{Heartbeats statistics, classes are divided according to \cite{liu2019arrhythmias}.
\label{tab:ECG-heartbeats-statistics}}
\end{table}

We perform data preprocessing steps similar to those in \cite{liu2019arrhythmias,lynn2019deep}.
First, we manually pick 38 subjects whose recordings have both MLII
and V1 channels and contain no paced beats. For each univariate signal,
the raw ECG signal is detrended by first fitting a 6-order polynomial
and then subtracting it from the signal. Following this, a 6-order
Butterworth bandpass filter with 5Hz and 15Hz range is applied. Finally,
filtered signals are normalised individually by using Z-score normalisation.
Each heartbeat is then represented by a window of length 360 samples,
with the R-peak values given in expert annotation. Each heartbeat
therefore corresponds to one second in data recordings.

In this experiment, we remove the unknown class and consider heartbeat
in one of three classes V, S, F as anomalous instances. We randomly
select 20\% of subjects for testing and the remaining subjects are
used for training. We perform 4-fold cross validation to optimise
the hyperparameters. The final AUC and F1 measures are calculated
from 3 different runs. 

\subsubsection*{Matrix construction}

The common practice is that each heartbeat is classified into one
of the predefined classes. We consider a different setting in which
we consider multiple consecutive heartbeats as one unit. The units
are extracted from original data using a sliding-window strategy.
Each unit is labelled abnormal if one of the heartbeats is abnormal.
Thus the data will become more balanced, as seen in Table~\ref{tab:ECG-statistics-prediction-lengths}.
The detection may be more sensitive as a result, allowing a way of
screening before the doctor has a detailed investigation into the
suspicious units. Another motivation is that for heartbeat prediction
models, predicting multiple beats at once may be easier than predicting
a sequence of beats due to their local dependencies.

The matrices are then constructed by using one heartbeat per row.
In essence, we are modelling the dependencies between wave signals
within a beat, and across multiple beats. This construction allows
using longer contexts. For example, when $N=10$ beats are grouped,
a sequential model of 20 steps will account for 200 beats, which is
much more difficult to handle by conventional beat-based models. 

\begin{table}[h]
\centering{}%
\begin{tabular*}{0.95\columnwidth}{@{\extracolsep{\fill}}|>{\centering}m{0.2\columnwidth}|c|>{\centering}m{0.12\columnwidth}|>{\centering}m{0.15\columnwidth}|>{\centering}m{0.13\columnwidth}|}
\hline 
\# prediction steps & Normal & Abnormal & \# test samples & \% test outlier \tabularnewline
\hline 
\hline 
5 heartbeats & 12,437 & 5,396 & 4,128 & 22\tabularnewline
\hline 
10 heartbeats & 5,334 & 3,398 & 2,026 & 29\tabularnewline
\hline 
20 heartbeats & 2,157 & 2,027 & 974 & 39\tabularnewline
\hline 
\end{tabular*}\caption{ECG statistics for different prediction lengths.\label{tab:ECG-statistics-prediction-lengths}}
\end{table}

\subsubsection*{Evaluation result}

Table~\ref{tab:ECG:-Performance-5-beats} compares the performances
of LSTM and matLSTM in predicting 5 heartbeats ahead. For LSTM models,
we adopt two different types of inputs, the first type uses one heartbeat
as the observation at each timestep and the second type uses flattened
vector of 5 heartbeats as the observation at each timestep. For matLSTM
model, a group of 5 heartbeats can be represented by a matrix, as
denoted above. From table~\ref{tab:ECG:-Performance-5-beats}, we
observe that matLSTM yields the best performance, compared to LSTM
models. For LSTM models using one heartbeat as input at each step,
the results suggest that using longer context helps improve the performance.

\begin{table}[h]
\begin{centering}
\begin{tabular*}{0.95\columnwidth}{@{\extracolsep{\fill}}|c|c|c|c|c|}
\hline 
Models & $T_{e}$ & \#Params & AUC (\%) & F1 (\%)\tabularnewline
\hline 
\hline 
\multicolumn{5}{|l|}{\emph{Input: 1 heartbeat at each timestep}}\tabularnewline
\hline 
LSTM & 10 & 386k & 90.7 $\pm$ 0.3 & 71.2 $\pm$ 0.6\tabularnewline
\hline 
LSTM & 45 & 822k & 91.2 $\pm$ 0.3 & 71.9 $\pm$ 0.9\tabularnewline
\hline 
\multicolumn{5}{|l|}{\emph{Input: 5 heartbeats at each timestep}}\tabularnewline
\hline 
LSTM-flat & 9 & 727k & 87.8 $\pm$ 0.2 & 69.9 $\pm$ 0.2\tabularnewline
\hline 
matLSTM & 9 & 257k & \textbf{92.5 $\pm$ 0.1} & \textbf{72.8 $\pm$ 0.2}\tabularnewline
\hline 
\end{tabular*}
\par\end{centering}
\caption{ECG: Performance of different models for predicting 5 heartbeats.
$T_{e}$: past context length. All models use the encoder-predictor
version. LSTM-flat denotes LSTM model using flattened vectors as inputs.
\label{tab:ECG:-Performance-5-beats}}
\end{table}

Table~\ref{tab:ECG:-Changes-in-length} reports the performances
of models under various context length and group size. We use one
heartbeat as input at each timestep for LSTM model since it gives
better performance than using a flattened vector of multiple heartbeats.
For fair comparisons, LSTM and matLSTM are compared using the same
context length. In every case, matrix models show better performance
than vector LSTM models and 3D-CNN models.

\begin{table}[h]
\begin{centering}
\begin{tabular*}{0.95\columnwidth}{@{\extracolsep{\fill}}|c|c|c|c|c|}
\hline 
Models & $T_{e}$ & \#Params & AUC (\%) & F1 (\%)\tabularnewline
\hline 
\hline 
\multicolumn{5}{|l|}{\emph{Predict 5 heartbeats ahead}}\tabularnewline
\hline 
3D-Conv-Predictor & 9 & 81k & 91.7 $\pm$ 0.1 & 71.4 $\pm$ 0.7\tabularnewline
\hline 
1-layer LSTM & 45 & 822k & 91.2 $\pm$ 0.3 & 71.9 $\pm$ 0.9\tabularnewline
\hline 
2-layer LSTM & 45 & 984k & 90.9 $\pm$ 0.2 & 71.9 $\pm$ 0.4\tabularnewline
\hline 
1-layer matLSTM & 9 & 257k & \textbf{92.5 $\pm$ 0.1} & \textbf{72.8 $\pm$ 0.2}\tabularnewline
\hline 
2-layer matLSTM & 9 & 343k & \textbf{92.5 $\pm$ 0.1} & \textbf{72.9 $\pm$ 0.4}\tabularnewline
\hline 
\multicolumn{5}{|l|}{\emph{Predict 10 heartbeats ahead}}\tabularnewline
\hline 
3D-Conv-Predictor & 9 & 106k & 90.9 $\pm$ 0.1 & 72.9 $\pm$ 0.2\tabularnewline
\hline 
1-layer LSTM & 90 & 1,010k & 89.2 $\pm$ 0.1 & 75.1 $\pm$ 0.2\tabularnewline
\hline 
2-layer LSTM & 90 & 1,243k & 89.1 $\pm$ 0.2 & \textbf{75.2 $\pm$ 0.4}\tabularnewline
\hline 
1-layer matLSTM & 9 & 263k & \textbf{91.4 $\pm$ 0.1} & 75.0 $\pm$ 0.2\tabularnewline
\hline 
2-layer matLSTM & 9 & 350k & \textbf{91.3 $\pm$ 0.1} & 74.7 $\pm$ 0.2\tabularnewline
\hline 
\multicolumn{5}{|l|}{\emph{Predict 20 heartbeats ahead}}\tabularnewline
\hline 
3D-Conv-Predictor & 9 & 143k & 90.4 $\pm$ 0.1 & 78.7 $\pm$ 0.2\tabularnewline
\hline 
1-layer LSTM & 180 & 1,308k & 87.3 $\pm$ 0.4 & 77.0 $\pm$ 0.2\tabularnewline
\hline 
2-layer LSTM & 180 & 1,670k & 87.0 $\pm$ 0.4 & 77.5 $\pm$ 0.7\tabularnewline
\hline 
1-layer matLSTM & 9 & 276k & \textbf{90.8 $\pm$ 0.1} & \textbf{79.7 $\pm$ 0.2}\tabularnewline
\hline 
2-layer matLSTM & 9 & 362k & \textbf{90.9 $\pm$ 0.1} & \textbf{79.9 $\pm$ 0.1}\tabularnewline
\hline 
\end{tabular*}
\par\end{centering}
\caption{ECG: Changes in performance with different prediction length. $T_{e}$:
past context length. LSTM denotes LSTM-Encoder-Predictor, matLSTM
denotes matLSTM-Encoder-Predictor.  \label{tab:ECG:-Changes-in-length}}
\end{table}

\selectlanguage{american}%

\section{Conclusion}

\selectlanguage{english}%
We have studied the task of unsupervised anomaly detection on temporal
multiway data. Unlike well-studied spatio-temporal data that exhibit
translation-invariance in time and space, multiway data are usually
\emph{permutation-invariant} in each of the modes. We investigated
the power and behaviours of matrix recurrent neural networks for the
task. Models were evaluated using a comprehensive suite of experiments
designed to expose model behaviours on synthetic sequences, moving
digits, and ECG recordings. Overall we empirically found that matrix
LSTMs, configured to predict the future subsequences, are highly suitable
for the problem. The models require a far less number of parameters
compared to the vector counterparts while capturing the temporal regularities
and predicting future well. The autoencoder configuration of the matrix
LSTMs, however, is not suitable for noisy data because of its high
memory capacity to compress the entire input sequence including the
noise. We also discovered a nice unintended consequence of matrix
RNNs: we can model accurately a very long sequence of vectors just
by rearranging data blocks into matrices.

\selectlanguage{american}%

\bibliographystyle{plain}
\bibliography{anomaly,../../bibs/truyen,../../bibs/ME}

\end{document}